# Virtual Reality Simulation of Fire Fighting Robot Dynamic and Motion

**Joga D. Setiawan\***, **Mochamad Subchan\***, and **Agus Budiyono** ⁼

\*Mechanical Engineering Department  
Diponegoro University, Semarang, Indonesia.  
e-mail: joga@mesin.ft.undip.ac.id

⁼Aeronautics and Astronautics Department  
Bandung Institute of Technology, Bandung, Indonesia.  
e-mail: agus.budiyono@ae.itb.ac.id

**Abstract**

This paper presents one approach in designing a Fire Fighting Robot which has been contested annually in a robotic student competition in many countries following the rules initiated at the Trinity College. The approach makes use of computer simulation and animation in a virtual reality environment. In the simulation, the amount of time, starting from home until the flame is destroyed, can be confirmed. The efficacy of algorithms and parameter values employed can be easily evaluated. Rather than spending time building the real robot in a trial and error fashion, now students can explore more variation of algorithm, parameter and sensor-actuator configuration in the early stage of design. Besides providing additional excitement during learning process and enhancing students understanding to the engineering aspects of the design, this approach could become a useful tool to increase the chance of winning the contest.

## 1  Introduction

Fire fighting robot (FFR) is an autonomous ground vehicle that has been popularly known to engineering students around the world. It has been contested annually in a robotic student competition in many countries following the rules initiated at the Trinity College, USA. The contest requires advanced mechatronics technology and knowledge using a handy robot as an educational tool [2].

The task of an FFR is to simulate a real-world operation of an autonomous robot performing a fire protection function in a real house. Starting from a home noted by "H" circle, an FFR has to find its way through an arena that represents a model house, find a lit candle that represents a fire in the house, extinguish the fire in the shortest time, and return to its home within a specified time.

This paper presents one approach in designing an FFR using computer animation in a virtual reality environment including one configuration example that consists of the mechanical design of the vehicle, the choice and arrangement of sensors and actuators, and the artificial intelligence of its controller.

The FFR has been developed to meet contest rules in [2]. As shown in Fig. 1, it is designed as a tracked vehicle with differential drive controlled by a unique algorithm embedded in its microcontroller. The control system the FFR shown in Fig. 2 will be mathematically modeled including its environment, which is the arena used in the competition shown in Fig 3.

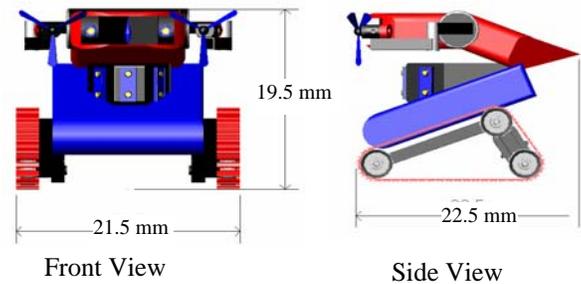

Front View    Side View

**Figure 1:** Fire Fighting Robot as a Tracked Vehicle

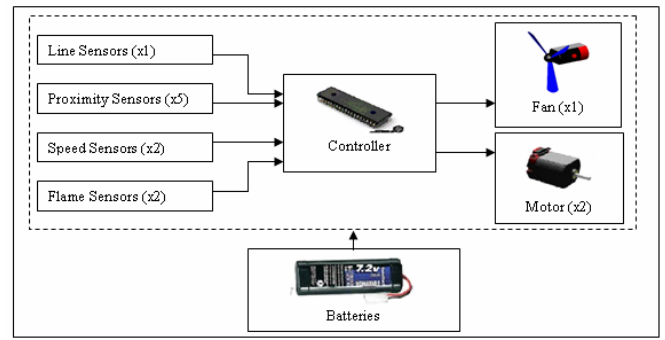

**Figure 2:** Control System of Fire Fighting Robot

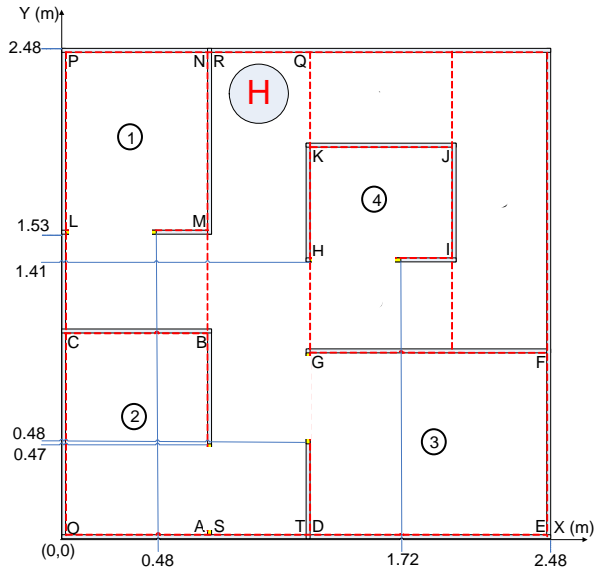

**Figure 3:** Contest Arena of Fire Fighting Robot

## 2 Mathematical Model

The FFR moving autonomously in the contest arena can be mathematically modeled according to three main groups: environment, kinematics and dynamics.

**2.1 Environment**
The mathematical model for the environment is built by determining the coordinates of walls that look like a labyrinth having four separated rooms as shown in Fig 3. All walls are assumed to have the same thickness.

According to the contest rules [1], white lines are available at the doorway of each room such that they can be used by FFR to determine whether it has moved to a room. In addition, white circumferential lines on the floor are provided around the home "H" and the only-one targeted candle. Thus, the coordinates of these circles are also noted since they can be used by FFR to determine whether to stop or not knowing it is at home or near a target.

The coordinates for the candle's location are set to be varied such that the candle can be anywhere in the four rooms. However, some contest rules are applied. For example, the candle will not be placed in a hallway, but it might be placed just inside a doorway of a room. The candle circle will not touch the doorway line, at least 33 cm into the room before it encounters the candle [1].

**2.2 Kinematics**
Fire Fighting Robot moves on X-Y plane with the velocity vector $V = [u\ v\ 0]^T$ at its center of mass (COM) shown in Fig. 4, where $u$ is the longitudinal speed and $v$ is the lateral speed. Since the motion is nonholonomic, $v=0$. However, it may rotate with an angular speed $\omega = [0\ 0\ r]^T$. If $q = [X\ Y\ \theta]^T$ is the state vector representing robot's X-Y position measured at the COM relative to the origin, where $\theta$ is robot's orientation, thus $\dot{q} = [\dot{X}\ \dot{Y}\ \dot{\theta}]^T$ is its velocity vector.

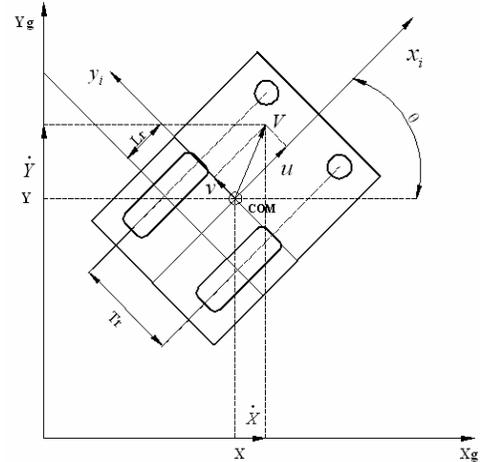

**Figure 4:** Robot's Position, Orientation and Geometry on X-Y Plane Coordinate System

The lateral and angular speeds, $u$ and $r$ can be determined by having angular speeds on the left and right drive wheels, $\omega_l$ and $\omega_r$

$$u = (\omega_r + \omega_l)\frac{R_t}{2} \quad (1)$$

$$\omega = r = (\omega_r - \omega_l)\frac{R_t}{T_r} \quad (2)$$

where $R_t$ is the radius of drive wheels and $T_r$ is the distance between the left and the right drive wheels. Therefore the velocity vector can be expressed as

$$\dot{q} = \begin{bmatrix}\dot{X}\\ \dot{Y}\\ \dot{\theta}\end{bmatrix} = \begin{bmatrix}\frac{1}{2}R_t\cos\theta & \frac{1}{2}R_t\cos\theta\\ \frac{1}{2}R_t\sin\theta & \frac{1}{2}R_t\sin\theta\\ \frac{R_t}{T_r} & -\frac{R_t}{T_r}\end{bmatrix}\begin{bmatrix}\omega_r\\ \omega_l\end{bmatrix} \quad (3)$$

**2.3 Dynamics**

**2.3.1 Tracked Vehicle:** The robot has one drive wheel and two sprockets on each side as shown in Fig 5. The normal force on the drive wheel noted as $N_c$ is

$$N_c = \frac{k}{2(L_r + k)}m.g \quad (4)$$

The friction force acting on the drive wheel $F_{si}$

$$F_{si} = \mu.N_c \quad (5)$$

where $\mu$ is the effective friction coefficient. Since the total moment requires to accelerate the wheels consisting one drive wheel, two sprockets and a belt is

$$\sum M_c = I_{wheels}\alpha\left(2 + \frac{L_2}{R_t}\right) \quad (6)$$

where $I_{wheels}$ is the effective wheel rotational inertia, thus the motor will need to overcome the torsional load

$$T_L = \sum M_c + F_{si}.R_t \qquad (7)$$

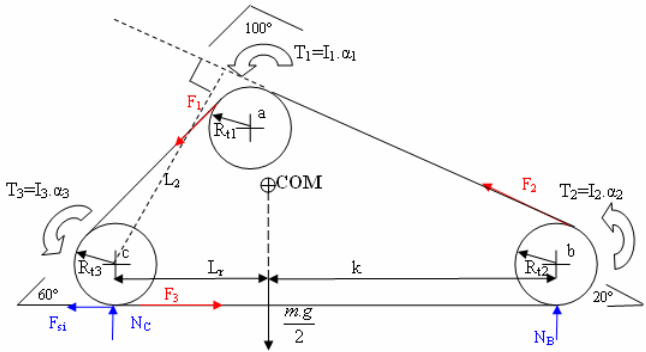

**Figure 5:** Forces and Torsion on One Side of Robot

**2.3.2 Sensors:** All sensors installed on the FFR as shown in Fig. 2 are assumed to have relatively very high bandwidth such that there is no need to model their dynamics. However, for the proximity sensors, the characteristic of GP2D12 infrared sensor is used [6]. Its calibration curve that relates the distance and the resulting voltage is incorporated to the simulation in order to reveal the effect of its nonlinear characteristic. The location of five proximities sensors from the top view of the FFR are shown in Fig 6. One proximity sensor is facing forward of the FFR noted as CF while four others are looking outward at each corner. Using simple geometric distance equations, the effect of the location and orientation of these sensors are included besides the effect of FFR location and orientation in order to navigate in the arena without hitting walls.

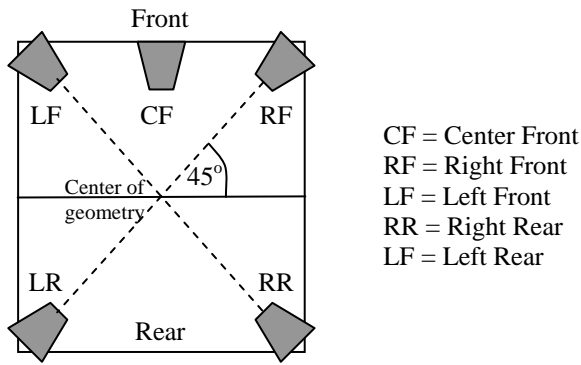

CF = Center Front
RF = Right Front
LF = Left Front
RR = Right Rear
LF = Left Rear

**Figure 6:** Location of Five Proximity Sensors

One line sensor is placed facing down to detect the white lines on the arena floor. The signal from this sensor can be modeled as a digital value 1 as soon as the sensor coordinate is at the location of white floors mentioned in section 2.1, otherwise the value is 0. An example of sensor used for this is a photo-reflector Hamamatsu P5587 [7].

Besides for detecting the existence of home and candle circles, the information form the white line sensors are useful for determining the zone or room of the FFR is currently located. This is done by having the controller to record the number of white line has been encountered.

The two flame sensors are installed at top center facing forward focusing at a specified distance from the FFR as shown in Fig. 7. FFR. The relative distance R of the candle flame is determined by the following equation

$$R = \sqrt{(X_C - X_R)^2 + (Y_C - Y_R)^2} \qquad (8)$$

where $(X_C, Y_C)$ is the candle location coordinate in m and $(X_R, Y_R)$ is the FFR location coordinate in m, in order to mathematically modeled the output signal in term of light intensity I in footcandle shown in Eq. 9.

$$I = \frac{0.0125}{R^2}K \quad , R > 0 \qquad (9)$$

where $K = f(d\theta)$ is the light intensity coefficient modeled in Fig. 8.

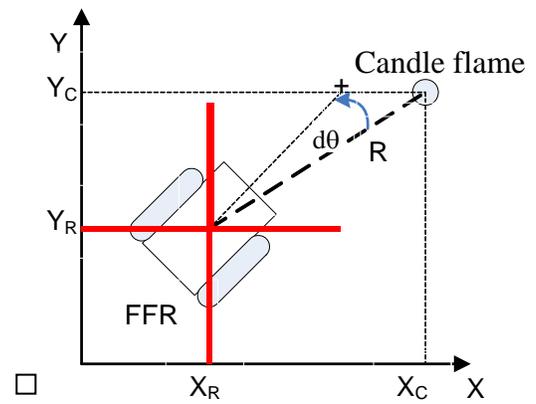

**Figure 7:** Relative Distance and Orientation of FFR for Flame Detection

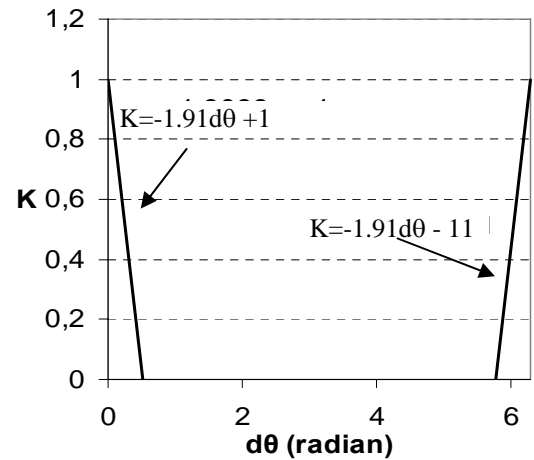

**Figure 8:** The Effect of FFR Relative Orientation to the Light Intensity Coefficient K

**2.3.3 Actuators:** Two equivalent DC motors are assumed to propel the robot. Each DC motor is connected to the drive wheel on each side. The motor generates torsion

$$T_m(s) = K_i I_a(s) \qquad (10)$$

where $K_i$ is torsion coefficient in N-m/A and $I_a$ is the armature current. This armature current depends on the applied voltage $E_a$ and back emf $E_b$ which are popularly known in many control textbooks [2]. For the sake of briefness, the discussion of the DC motor model is thus ommitted here, eventhough the model is included in the simulation.

Using Eqs. 6, 7 and 10, the dynamic equation of each wheel is therefore can summarized to

$$(J_m + I_{wheels})\left(2 + \frac{L_2}{R_t}\right)\frac{d\omega_m(t)}{dt} + B_m.\omega_m(t) = T_m(t) \qquad (11)$$

where the rotor inertia of the DC motor is represented by $J_m$, and other viscous friction on the motor bearing, gearbox and wheels bearings are lumped to $B_m$.

Other actuators are the two DC motors to drive two fans. These fans are turned-on only to the destroy the candle flame. It is assumed that after the FFR approached the flame and the fans are activated for 4 seconds, then the flame is put off. Thus, there is no need to model the dynamic of these fans. However, for animation purposes, the fans are shown to rotate during this action in the virtual reality environment.

**2.4 Controller**
The controller used is assumed to have high enough sampling rate relative to the traveling speed of the FFR such that the controller dynamic can be neglected. Moreover, the controller is modeled to have sufficient input and output channels that compatible with all sensors and actuators utilized by the FFR. Furthermore the controller is modeled to be capable of performing counting operation.

### 3 Navigation Algorithm

A navigation algorithm used for the *FFR* to move efficiently in the arena has been uniquely developed. The algorithm can be separated into three main tasks:
1. Navigation going through corridors
2. Navigation in rooms
3. Navigation to return home

**3.1 Navigation through Corridor**
There are three basic motions that the FFR can perform in this task [18]:
1. Moving forward that depends on the readings from the proximity sensors.
2. Turning that is determined by the odometry of encoders connected to the right and left drive wheels. The odometry reading is used to estimate how many degrees the FFR has rotated during turning. When to turn to the left or to the right depends on the six identified cases shown in Fig. 9.
3. Wall Following. This motion is needed to make sure the FFR is always moving in parallel with detected side walls as shown in Fig. 10.

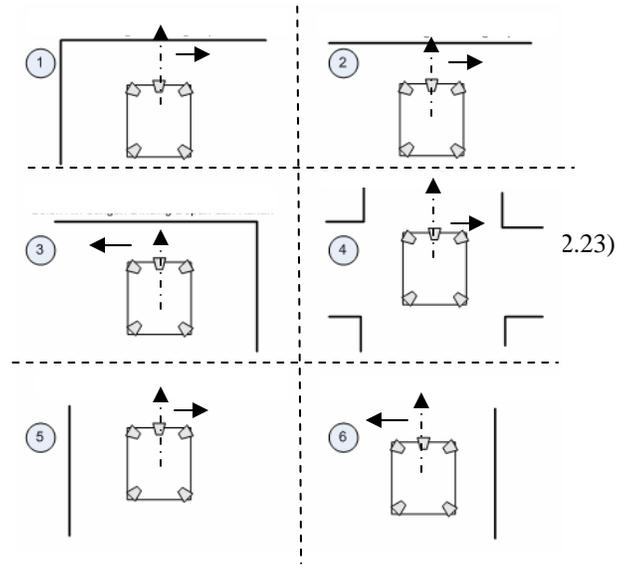

**Figure 9:** FFR Turns due to Encountered Wall Cases

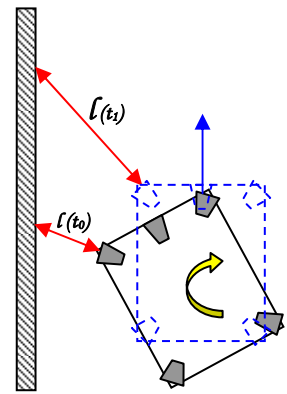

**Figure 10:** Wall Following using Information from at Least Two Proximity Sensors

**3.2 Nagivation in Rooms**
Once the FFR identifies that it has been in a room and had few distance passing a white line on a doorway, the controller will switch to the task noted as the Navigation in Rooms as shown in Fig 11.

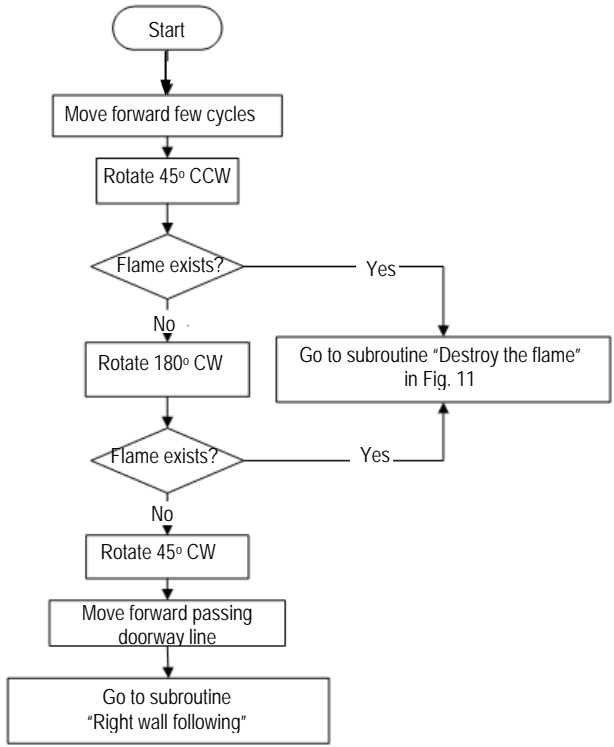

**Figure 11:** Task of Navigation in Rooms

### 3.3 Nagivation to Destroy the Flame
The FFR may detect existence of a flame during rotating and scanning a room and therefore the controller will switch the task to the Navigation to Destroy the Flame as shown in Fig. 12.

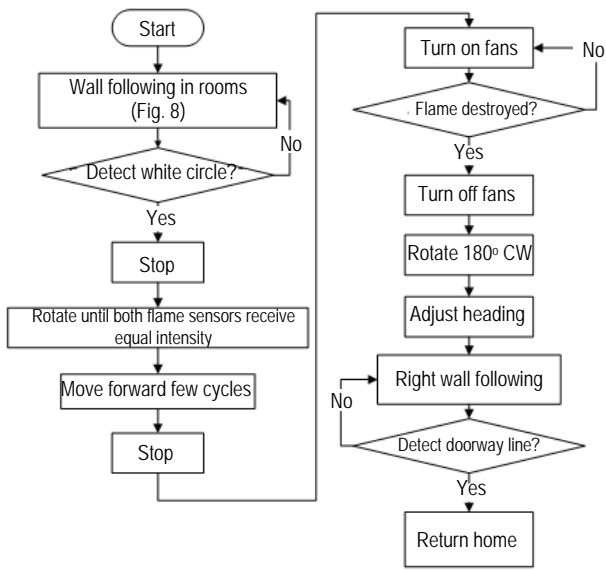

**Figure 12:** Task of Navigation to Destroy the Flame

### 3.4 Nagivation to Return Home
The controller will switch to the task Navigation to Return Home after the flame has been extinguished. This task is basically use Wall Following motion explained in section 3.1. At the end the FFR is set to stop after few distance passing the white circle of the Home.

## 4 Virtual Reality Simulation

Algorithm of FFR navigation is implemented on State Flow of MATLAB/Simulink[TM] from Mathworks as shown in Fig. 12. To travel along the corridor, the subsystem "To_Room" is used in which it has two subroutines: Wall Following and and Turn Program

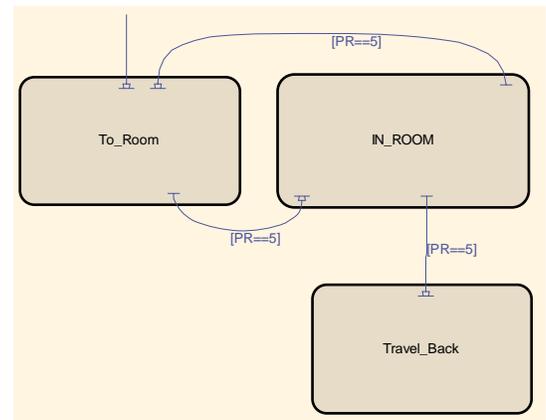

**Figure 13:** State Flow of Main Algorithm

Fire Fighting Robot virtual reality is created using Virtual Reality Toolbox in MATLAB/Simulink. The result of the virtual reality environment can be seen in Fig. 14. The FFR trajectories of four cases are provided in Fig. 15 and the elapsed times required to destroy the flame and the total time until the FFR returns home are shown in Table 1. The Simulink block diagram of the overall FFR mathematical model and algorithm is shown in Figure 16. Simulation parameters are obtained in [4]

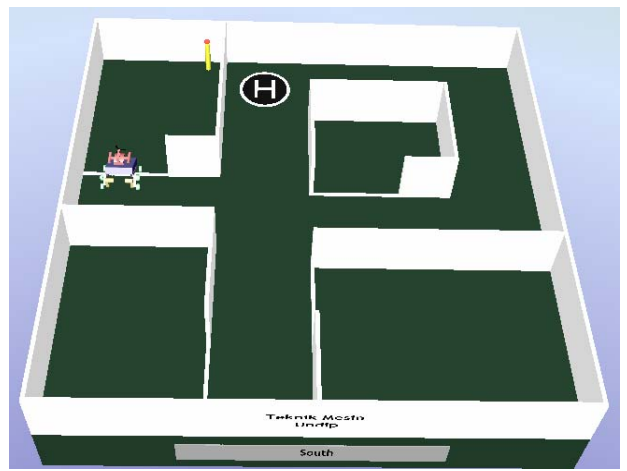

**Figure 14:** View of Virtual Reality Environment

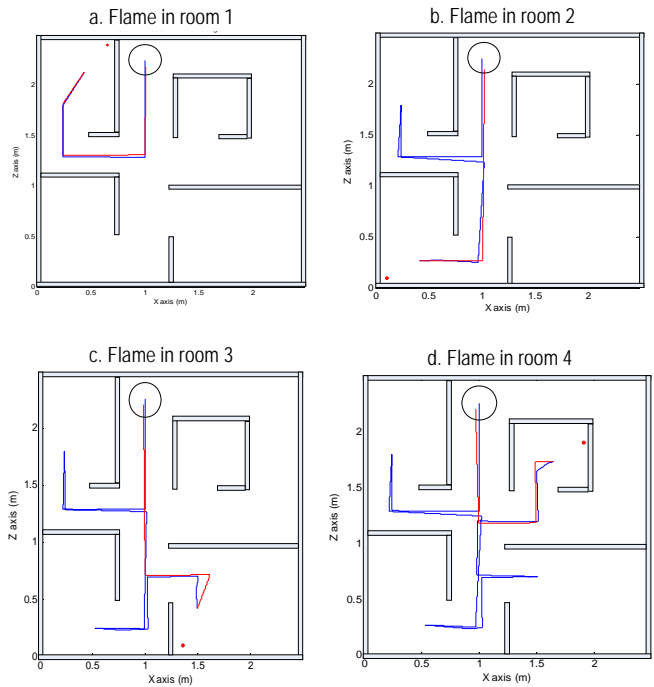

**Figure 15: Example of Four Trajectories of FFR**

**Table 1.** Example of four cases showing the elapsed time required to destroy the flame and return home

| Flame location | Room 1 | Room 2 | Room 3 | Room 4 |
|---|---|---|---|---|
| Time to destroy the flame (s) | 40.8 | 84.2 | 120.6 | 156.8 |
| Total time until return home (s) | 82.6 | 123.6 | 161.1 | 196.3 |

The animation results shown in Figure 15 reveal that the robot can successfully find the candle in four cases and navigate through the arena without hitting walls. The elapsed time performance of robot seen in Table 1 satisfies time limits of the contest regulation; maximum 5 minutes to find the candle and maximum 2 minute in return trip mode [2].

## 5 Conclusions

This paper describes the viability of simulation and animation of Fire Fighting Robot in order to evaluate the performance of the robot design in meeting some of the contest rules such as navigating in a labyrinth arena without hitting walls, quickly extinguishing a flame in a room and return home.

This work shows the benefit of virtual reality tool that enables students to quickly evaluate and clearly visualize the dynamic and motion of Fire Fightting Robot and the interaction between the robot and its environment before spending too much time in building the robot. The present model gives a reasonably accurate analytical representation of an example Fire Fighting Robot and its contest arena. In turn, students can become more familiar with the analytical models. This model can give the basis of a model that could be used by students to explore more innovative design for their robot.

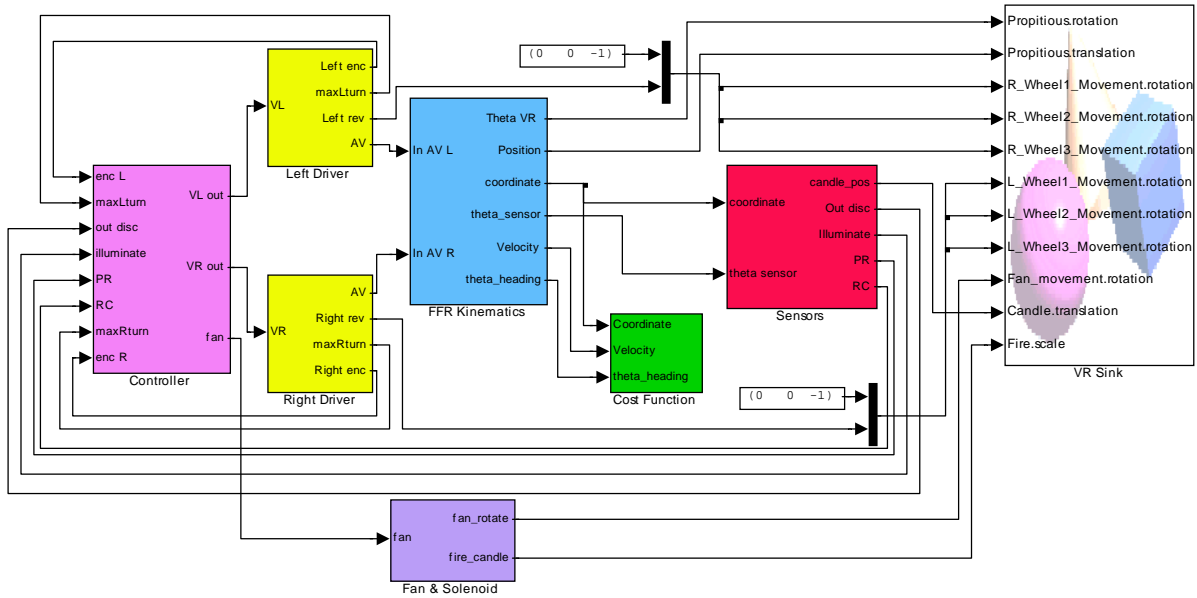

**Figure 16:    Simulink Block Diagram of FFR Model and Algorithm**